%% file: main.tex
\documentclass{article}

\usepackage[square,numbers]{natbib}
\usepackage[preprint]{neurips_2021}

\usepackage{graphicx}
\usepackage{graphics}
\usepackage[utf8]{inputenc} % allow utf-8 input
\usepackage[T1]{fontenc}    % use 8-bit T1 fonts
\usepackage[pagebackref=true,breaklinks=true,letterpaper=true,colorlinks,bookmarks=false,citecolor=cyan]{hyperref}
\usepackage{url}            % simple URL typesetting
\usepackage{booktabs}       % professional-quality tables
\usepackage{amsfonts}       % blackboard math symbols
\usepackage{nicefrac}       % compact symbols for 1/2, etc.
\usepackage{microtype}      % microtypography
\usepackage{mathtools}
\usepackage{multirow}
\usepackage{subfigure}
\usepackage{tabularx}
\usepackage{amsfonts}
\usepackage{wrapfig,lipsum,booktabs}
\usepackage[dvipsnames]{xcolor}
\usepackage{hyperref}
\usepackage{floatflt}
\usepackage{wrapfig,lipsum,booktabs}
\usepackage{bbm}
\usepackage{amssymb}
\usepackage{pifont}
\usepackage{enumitem}

\newcommand{\ours}{TransDA~}
\title{Transformer-Based Source-Free Domain Adaptation}
\author{%
  Guanglei Yang$^1$ \, Hao Tang$^2$ \, Zhun Zhong$^2$ \, Mingli Ding$^1$ \, Ling Shao$^3$ \,  Nicu Sebe$^{2}$ \, Elisa Ricci$^{14}$\\
  $^1$Haerbin Institute of Technology \, $^2$University of Trento \, $^3$IIAI \, $^4$Fondazione Bruno Kessler
}

\def\eg{\textit{e.g.}}
\def\ie{\textit{i.e.}}

\begin{document}

\maketitle

\begin{abstract}
In this paper, we study the task of source-free domain adaptation (SFDA), where the source data are not available during target adaptation. Previous works on SFDA mainly focus on aligning the cross-domain distributions. However, they ignore the generalization ability of the pretrained source model, which largely influences the initial target outputs that are vital to the target adaptation stage. To address this, we make the interesting observation that the model accuracy is highly correlated with whether or not attention is focused on the objects in an image. To this end, we propose a generic and effective framework based on Transformer, named TransDA, for learning a generalized model for SFDA. Specifically, we apply the Transformer as the attention module and inject it into a convolutional network. By doing so, the model is encouraged to turn attention towards the object regions, which can effectively improve the model's generalization ability on the target domains. Moreover, a novel self-supervised knowledge distillation approach is proposed to adapt the Transformer with target pseudo-labels, thus further encouraging the network to focus on the object regions. Experiments on three domain adaptation tasks, including closed-set, partial-set, and open-set adaption, demonstrate that \ours can greatly improve the adaptation accuracy and produce state-of-the-art results.~The source code and trained models are available at~\url{https://github.com/ygjwd12345/TransDA}.
\end{abstract}

\input{tex/intro.tex}

\input{tex/2related}
\input{tex/method.tex}

\input{tex/experiments}

\input{tex/conclusion}

\clearpage

\small
\bibliographystyle{plainnat}
\bibliography{nips}

\end{document}

%% file: tex/intro.tex
\section{Introduction}
Deep learning has enabled several advances in various tasks in computer vision, such as image classification~\cite{deng2009imagenet,he2016deep}, object detection~\cite{girshick2015fast,ren2015faster}, semantic segmentation~\cite{long2015fully}, and so on.
However, deep models suffer from significant performance degradation when applied to an unseen target domain, due to the well-documented domain shift problem~\cite{glorot2011domain}. 
To solve this problem, domain adaptation has been introduced, aiming to transfer knowledge from a fully labeled source domain to a target domain~\cite{long2015learning,tzeng2017adversarial,zhang2017curriculum}.
A common strategy in domain adaptation is to align the feature distribution between the source and target domains by minimizing the domain shift through various metrics, such as Correlation Distances~\cite{yao2015semi,sun2016return}, Maximum Mean Discrepancy~\cite{long2016unsupervised,long2017deep,kang2019contrastive}, Sliced Wasserstein Discrepancy~\cite{lee2019sliced}, and Enhanced Transport Distance~\cite{li2020enhanced}. 
Another popular paradigm leverages the idea of adversarial learning~\cite{goodfellow2014generative} to minimize cross-domain discrepancy~\cite{yang2020bi,li2020model,sankaranarayanan2018generate,saito2018maximum,kurmi2019attending}.

Despite the success of current domain adaptation methods, they work under the strict condition that the source data are always available during training. 
However, such a condition has two drawbacks that hinder the application of these methods. First, the source datasets, such as VisDA~\cite{peng2017visda} and GTAV~\cite{richter2016playing}, are usually large and imply high saving and loading costs, restricting their use on certain platforms, especially portable devices. Second, fully accessing the source data may violate data privacy policies. To avoid this issue, companies or organizations prefer to provide the learned models rather than the data. Therefore, designing a domain adaptation method without requiring the source datasets has great practical value. 
To this end, in this paper, we aim to address the recently introduced problem of source-free domain adaptation~\cite{liang2020we} (SFDA), in which only the model pretrained on the source and the unlabeled target dataset are provided for target adaptation.

\begin{figure*}[!t] \small
    \centering
    \includegraphics[width=1\linewidth]{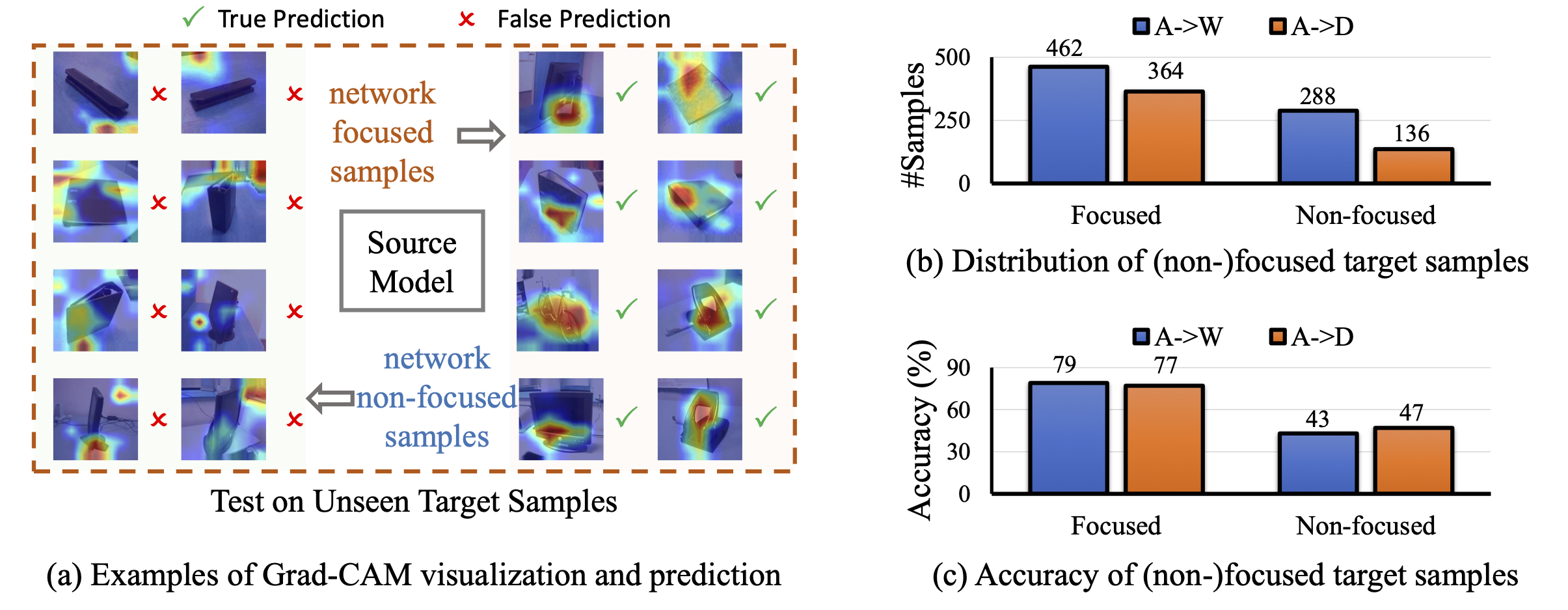}
    \vspace{-.12in}
    \caption{Evaluation of attention for the source model on Office-31 dataset. (a) Examples of Grad-CAM~\cite{jacobgilpytorchcam} visualization and prediction results in A$\to$W. (b) The distribution and (c) the accuracy of the source model on focused and non-focused samples in A$\to$W and A$\to$D.}
    \vspace{-.12in}
    \label{fig:teaser}
\end{figure*}

So far, few works~\cite{li2020model,liang2020we,ahmed2021unsupervised} has been proposed for SFDA, aiming to align the source and target distributions by learning with underlying constraints, such as information maximization and label consistency. 
However, all of these perform adaptation with a model pretrained on the source data, while neglecting the source model's generalization ability. In SFDA, the adaptation process largely relies on the accuracy of the source model on the target domain.
Without a source model that generalizes well, the generated pseudo-labels may contain significant noise, and learning with them will undoubtedly harm the model's performance. In this paper, we attempt to improve the generalization ability of the source model for SFDA. Different from existing out-of-domain generalization methods~\cite{qiao2020learning,qiao2021uncertainty}, which aims to improve the model generalization by augmenting the diversity of the source samples, in this paper we introduce a new perspective for building a robust source model, motivated by the following observation. In Figure~\ref{fig:teaser}, we directly apply the source model on the unseen target samples (A$\rightarrow$W and A$\rightarrow$D on Office-31~\cite{saenko2010adapting}) and produce the heat maps by Grad-CAM~\cite{jacobgilpytorchcam}.
We use Amazon Mechanical Turk and ask annotators to label the samples with ``focused / non-focused'' according to whether the heat map (red region) is localized on the object. Examples are shown in Figure~\ref{fig:teaser}(a). We observe that the accuracy of the focused samples is much higher than that of the non-focused samples (see Figure~\ref{fig:teaser}(b, c)). This finding reveals that if a network can effectively focus on the objects in the images, it will have a high prediction accuracy on these images. 

Inspired by the above observation, we propose \ours for SDFA by equipping a convolutional model with a Transformer~\cite{dosovitskiy2020image} module, which can effectively encourage the network to focus on the objects and thus improve the performance on target samples. 
Specifically, by injecting the Transformer after the last convolutional layer of ResNet-50, we can leverage its long-range dependence to force the model to pay more attention to the objects. 
Albeit simple, this modification can significantly improve the generalization ability of the source model. In addition, during the target adaptation, we propose a self-supervised knowledge distillation with generated pseudo-labels, further leading the Transformer to learn to focus on the objects of target samples. We evaluate our \ours on three domain adaptation tasks, including closed-set~\cite{saenko2010adapting}, partial-set~\cite{cao2018partial}, and open-set~\cite{panareda2017open} domain adaptation. Extensive results demonstrate that \ours outperforms state-of-the-art methods on all tasks. For example, on the Office-Home dataset, \ours advances the best accuracies from 71.8\% to \textbf{79.3\%}, from 79.3\% to \textbf{81.3\%}, and from 72.8\% to \textbf{76.8\%} for the closed-set, partial-set and open-set settings, respectively. To summarize, this work provides the following three contributions:
\vspace{-.1in}
\begin{itemize}[leftmargin=0.15in]
    \item Through an in-depth empirical study, we reveal for the first time the importance of the network attention for SFDA. %\eli{We reveal, through an empirical study, that the ability of the source network to focus on objects'regions is a crucial factor for out-of-domain generalization.}
    This provides a new perspective for improving the generalization ability of the source model.
     \vspace{-.04in}
    \item We propose a Transformer-based network for SFDA, which can effectively lead the model to pay attention to the objects, and thus significantly increases the model generalization ability. To our knowledge, we are the first to propose Transformer for solving the domain adaptation task.
     \vspace{-.04in}
    \item We introduce a novel self-supervised knowledge distillation approach to further help the Transformer to focus on target objects.

\end{itemize}

%% file: tex/2related.tex
\section{Related Work}
\vspace{-.1in}

\textbf{Traditional Domain Adaptation.} Domain adaptation aims to improve the target learning by using a labeled source domain that belongs to a different distribution.
%, whose distribution is different from the target domain. 
%Spurred by the recent advances in deep convolutional neural networks, 
A number of methods have been proposed for unsupervised domain adaptation. One common solution is to guide the model to learn a domain-invariant representation by minimizing the domain discrepancy~\cite{yao2015semi,sun2016return,long2016unsupervised,long2017deep,kang2019contrastive}. For example, ETD~\cite{li2020enhanced} employs an enhanced transport distance to reflect the discriminative information. 
In a similar way, CAN \cite{kang2019contrastive} optimizes the network by considering the discrepancy between the intra- and the inter-class domains.
More recently, several methods have focused on the feature discrimination ability during domain adaptation. They introduce different normalization or penalization strategies to boost the feature discrimination ability on the target domain, such as batch normalization \cite{carlucci2020multidial}, batch spectral penalization~\cite{chen2019transferability}, batch nuclear-norm maximization~\cite{cui2020towards}, and transferable normalization~\cite{Wang19TransNorm}.
Another branch of methods exploit a generative adversarial network to address the domain confusion \cite{ganin2016domain,tzeng2017adversarial}. BDG~\cite{yang2020bi} and GVB-GD~\cite{cui2020gradually} use bi-directional generation to construct domain-invariant representations. Despite the large success of the above methods, they typically require access to source data when learning from the target domain. This may lead to personal information leakage and thus violate data privacy requirements.
%and go against the current trend of decentralization.

\textbf{Source-Free Domain Adaptation}. To alleviate the issue of data privacy, recent works~\cite{li2020model,liang2020we,ahmed2021unsupervised} have turned their attention to the problem of source-free domain adaptation, where only the pretrained source model and the target samples are provided during the target adaptation stage. 
C3-GAN~\cite{li2020model} introduces a collaborative class conditional generative adversarial net to produce target-style training samples. In addition, a clustering-based regularization is also used to obtain more discriminative features in the target domain. Meanwhile, SHOT~\cite{liang2020we} freezes the classifier module and learns the target-specific feature extractor with information maximization and self-supervised pseudo-labeling, which can align the target feature distribution with the source domain. DECISION~\cite{ahmed2021unsupervised} extends SHOT to a multi-source setting by learning different weights for each source model. 
Different from the above methods, in this paper, we attempt to improve the generalization ability of the source model by injecting a Transformer module into the network. Our approach can be readily incorporated into most existing frameworks to boost the adaptation accuracy.

\textbf{Vision Transformers}. The Transformer was first proposed by \cite{vaswani2017attention} for machine translation
and has been used to establish state-of-the-art results in many natural language processing tasks. 
Recently, the Vision Transformer (ViT) \cite{dosovitskiy2020image} achieved state-of-the-art results on the image classification task by directly applying Transformers with global self-attention on full-sized images. Since then, Transformer-based approaches have been shown to be efficient in many computer vision tasks, including object detection \cite{carion2020end,zhu2020deformable}, image segmentation \cite{chen2021transunet,zheng2020rethinking}, video inpainting \cite{zeng2020learning}, virtual try-on \cite{ren2021cloth}, video classification \cite{neimark2021video}, pose estimation \cite{huang2020hand,huang2020hot}, object re-identification \cite{he2021transreid}, image retrieval \cite{el2021training}, and depth estimation \cite{yang2021transformers}.
Different from these approaches, in this paper, we adopt a Transformer-based network to tackle the source-free domain adaptation task.
To this end, we propose a generic yet straightforward TransDA framework for domain adaptation to encourage the model to focus on the object regions, leading to improved domain generalization.

%% file: tex/method.tex
\section{Method}\label{sec:method}
\begin{figure*}[t] \small
    \centering
    \includegraphics[width=1\linewidth]{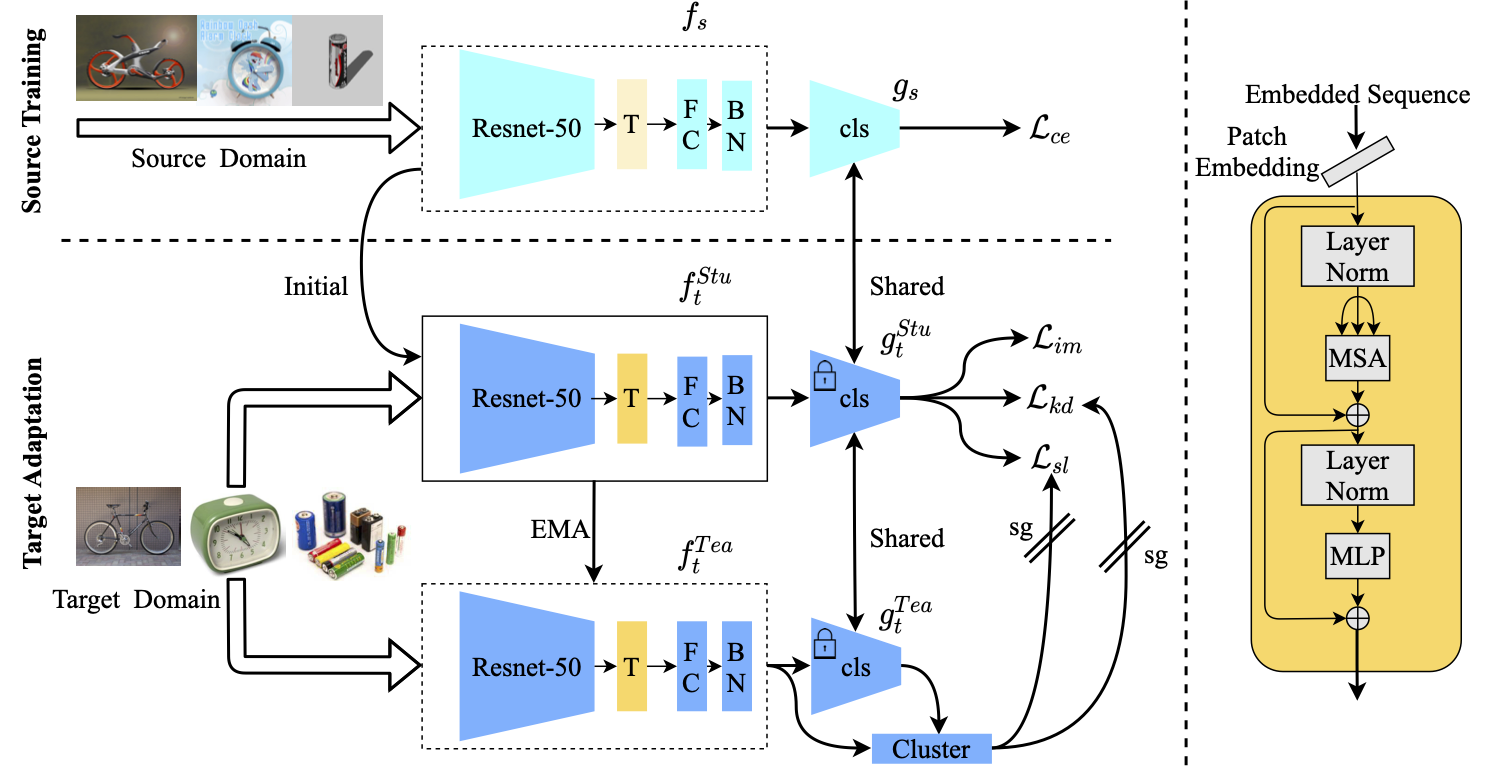}
    \caption{Overview of the proposed method. The model includes a feature extractor $f$ and a classifier $g$. We inject a Transformer module (structure is shown on the right) into $f$ to obtain a representation with improved generalization capability. First, we train the model with the labeled source data. Then, we create a teacher model ($Tea$) and a student ($Stu$), cloned from the source model, for target adaptation. The teacher model is used to produce pseudo-labels for computing the self-labeling loss ($\mathcal{L}_{sl}$) and knowledge distillation loss ($\mathcal{L}_{kd}$). We also calculate the information maximization loss ($\mathcal{L}_{im}$) based on the outputs of the student model. Here, \textit{sg}: stop-gradient, \textit{EMA}: exponential moving average, \textit{FC}: fully connected layer, \textit{BN}: batch normalization layer.}
    \label{fig:overview}
    \vspace{-0.4cm}
\end{figure*}

\subsection{Problem Definition}

In traditional domain adaptation (DA), models are trained on an unlabeled target domain $X_{t} {=}  \{x_{t}^i\}^{M}_{i=1}$ and a source domain $X_{s} {=} \{x_{s}^j\}^{N}_{j=1}$ along with corresponding labels $Y_{s} {=} \{y_{s}^j\}^{N}_{j=1}$, where $y_s^j$ belongs to the set of $K$ classes. The distributions of source and target data are denoted as $x_{s}{\sim} p_{data}(x_{s})$ and $x_{t}{\sim} p_{data}(x_{t})$, respectively, where $ p_{data}(x_{t}) {\neq} p_{data}(x_{s})$. The goal is to learn a target network $\phi_t$, using the labeled source data and unlabeled target data, that can accurately recognize the target samples. In SFDA, (1) the labeled source data is only used to pretrain a source model $\phi_s$; and (2) the target network $\phi_t$ is learned with the pretrained source model $\phi_s$ and the unlabeled target data $X_{t}$.

\subsection{Overview}

The overall framework of the proposed method, which consists of (1) source training and (2) target adaptation, is shown in Figure~\ref{fig:overview}. First, we train the source model $\phi_s$ with samples from the source domain using the cross-entropy loss. The source model $\phi_s$ is composed  of two modules: the feature extractor $f_s{:}\mathcal{X}{\to}\mathcal{R}^d$ and the classifier $g_s{:}\mathcal{R}^d{\to}\mathcal{R}^K$, where $d$ indicates the dimension of the output of the feature extractor and $K$ refers to the number of categories in the source data. Therefore, we have $\phi_s{=}f_s{\circ}g_s$. Different from existing methods that use a convolutional neural network (CNN) as the feature extractor (\eg, ResNet~\cite{he2016deep}), we propose to inject a Transformer module~\cite{dosovitskiy2020image} after the last layer of the CNN, which can help the feature extractor pay more attention to the objects and thus produce a more robust representation. 
Second, we aim to learn a network $\phi_t$ in the target domain, given the pretrained $\phi_s$. Specifically, we maintain a teacher model $\phi_t^{Tea}$ and a student model $\phi_t^{Stu}$, which are both initialized by the parameters of $\phi_s$. Following~\cite{liang2020we}, we fix the classifiers and only update the feature extractors. The updating strategies for the feature extractors are different. For $f_t^{Stu}$, we update it with the gradients produced by the target adaptation losses. $f_t^{Tea}$ is updated with an exponential moving average of the parameters of $f_t^{Stu}$. The teacher model $\phi_t^{Tea}$ is used to produce a hard pseudo-label $\hat{y}_t$ and soft pseudo-label $\bar{y}_t$ for calculating the self-labeling loss and knowledge distillation loss, respectively. In addition, the information maximization loss is computed with the outputs of $\phi_t^{Stu}$. The above three losses are used to update $f_t^{Stu}$, where the information maximization and self-labeling losses are employed to align the cross-domain feature distributions and the knowledge distillation loss is designed to force the model to focus on objects.

\subsection{Model Pretraining on Source with Transformer}

\textbf{Loss Function}. In this stage, we aim to learn a source model with the labeled source data. Generally, given a network $\phi_s$ initialized on ImageNet~\cite{deng2009imagenet}, we train it with the label smoothing cross-entropy loss~\cite{muller2019does}:
\begin{equation}\label{eq:featureconditionalkernel}
\begin{aligned}
\mathcal{L}_{ce}=-\mathbb{E}_{x_s\in X_s}\sum_{k=1}^K \hat{y}_k\log{\sigma(\phi_s(x_s))},
\end{aligned}  
\end{equation}
where $\hat{y}_k{=}(1-\alpha)~y_k{+}\alpha/K$. $\alpha$ is the smoothing factor and is empirically set to 0.1. $\sigma(.)$ is the softmax operation.

\textbf{Discussion}. In SFDA, the target adaptation stage greatly relies on the target outputs obtained by the source model. Hence, it is important to learn a robust source model, such as to counteract with domain bias. As shown in Figure~\ref{fig:teaser}, the model can effectively classify the target samples if it can localize the object regions. This observation is reasonable, because the model learns to capture the common object patterns instead of domain-specific information (\eg, background, color distribution) if it can always focus on the objects. Therefore, one solution for improving the generalization ability is forcing the model to focus on objects during training. Existing approaches typically select a CNN model as the feature extractor of $\phi_s$. However, due to the intrinsic locality of the convolution operation (\ie,~small receptive fields), the CNN model prefers to capture local information, which may lead it to overfit on domain-specific information and thus fail to focus on objects, especially when encountering a large domain shift. %To the best of our knowledge, no previous works on DA have performed a similar analysis.

\textbf{Equipping the Model with a Transformer}. To address the drawback of CNNs in SFDA, in this paper, we propose to inject a Transformer module after the convolutional network~\cite{dosovitskiy2020image}. By doing so, we can leverage the property of the Transformer to capture long-range dependencies and explicitly encourage the model pay more attention to the objects. This enables us to reduce the impact of domain-specific information and produce more robust and transferable representations.

Specifically, we construct the feature extractor by injecting the Transformer after the last convolutional layer of ResNet-50~\cite{he2016deep}. Since the input of the Transformer should be a sequence, we first reshape the output of the ResNet-50 backbone $F{\in} \mathcal{R}^{h \times w \times \hat{d}}$ to  $\hat{F}{\in} \mathcal{R}^{u \times \hat{d}}$, where $h$, $w$, and $\hat{d}$ indicate the height, width, and dimension of $F$, respectively. $u$ is the product of $h$ and $w$. Then, $\hat{F}$ is regarded as the input sequence with patch embeddings for the Transformer.

In the first layer of the Transformer, we map the dimension of the patch embeddings $\hat{F}$ from $\hat{d}$ to $\bar{d}$ with a linear projection layer, producing $Z_0 {\in} \mathcal{R}^{u \times \bar{d}}$. Then, $Z_0$ is fed into $L$ Transformer layers, which include multi-headed self-attention (MSA) and multi-layer perceptron (MLP) blocks. 

Given the feature $Z_{l-1}$ obtained from the $l{-}1$-th Transformer layer, MSA ($Z_{l-1}$) is defined as:
\begin{equation}
    {\rm MSA}(Z_{l-1}) = Z_{l-1} + {\rm cont}({\rm AH}_1({\rm LN}(Z_{l-1})); {\rm AH}_2({\rm LN}(Z_{l-1})); \cdots; {\rm AH}_m({\rm LN}(Z_{l-1})))\times W,
\end{equation}
where AH represents a self-attention head~\cite{vaswani2017attention}, $\rm{cont}(\cdot))$ is the concentration operation,  ${\rm LN}(\cdot)$ is the layer normalization~\cite{wang2019learning}, and $m$ indicates the number of self-attention heads. $W{\in} \mathbb{R}^{m \cdot \check{d} \times \bar{d}}$ are the learnable weight matrices, where $\check{d}$ is the output dimension of each AH. 

The output of ${\rm MSA}$ is then transformed into an MLP block with a residual skip, formulated as:
\begin{equation}
    Z_{l}= {\rm MLP}({\rm LN}({\rm MSA}(Z_{l-1}))) + {\rm MSA}(Z_{l-1}).
\end{equation}
The structure of the Transformer layer is illustrated in the right part of Figure~\ref{fig:overview}. 

Given the output $Z_l {\in} \mathcal{R}^{u \times \bar{d}}$ of the Transformer, 
we obtain the global feature by average pooling, which is fed into the following layers, including one FC layer, one BN layer, and the classifier $g_s$.

\subsection{Self-Training on Target with Transformer}

\textbf{Information Maximization}. In the target adaptation stage, we are given a pretrained source model $\phi_s$ and the unlabeled target domain. We first initialize the target model $\phi_t$ with the parameters of $\phi_s$. Following \cite{liang2020we}, we fix the classifier $g_t$ to maintain the class distribution information of the source domain, and update the feature extractor $f_t$ using the information maximization (IM) loss~\cite{bridle1992unsupervised,gomes2010discriminative,hu2017learning}. This enables us to reduce the feature distribution gap between the source and target domains. The IM loss consists of a conditional entropy term and a diversity term:
\begin{equation}
    \mathcal{L}_{im}=-\mathbb{E}_{x_t\in X_t}\sum_{k=1}^K\sigma(\phi_t(x_t))\log \sigma(\phi_t(x_t))+\sum_{k=1}^K\bar{p}_k\log\bar{p}_k,
\end{equation}
where $\bar{p}{=}\mathbb{E}_{x_t\in X_t}[\sigma(\phi_t(x_t))]$ is the mean of the softmax outputs for the current batch. 

\textbf{Self-Labeling}. Although the IM loss can make the predictions on the target domain more confident and globally diverse, it is inevitably affected by the noise generated by wrong label matching. To address this issue, one solution is to utilize a self-labeling strategy to further constrain the model. In this paper, we use self-supervised clustering~\cite{liang2020we,ahmed2021unsupervised,li2020model} to generate pseudo-labels for target samples. Specifically, we first compute the centroid for each class in the target domain by weighted k-means clustering,
\begin{equation}\label{eq:initial center}
    \mu_k^{(0)}=\frac{\sum_{x_t\in X_t}\sigma(\phi_{t}(x_t))f_t(x_t)}{\sum_{x_t\in X_t}\sigma(\phi_{t}(x_t))}.
\end{equation}
Then, the initial pseudo-labels are generated by the nearest centroid classifier:
\begin{equation}\label{eq:initial pseudo lable}
    \hat{y}_t=\arg\min_k 1-\frac{f_t(x_t)\cdot \mu_k^{(0)}}{||f_t(x_t)||_2||\mu_k^{(0)}||_2},
\end{equation}
where $||*||_2$ denotes the $L2$-norm. Finally, the class centroids and pseudo-labels are updated as follows:
\begin{equation}\label{eq:loop}
        \mu_k^{(1)}=\frac{\sum_{x_t\in X_t}\xi(\hat{y}_t=k)f_t(x_t)}{\sum_{x_t\in X_t}\xi(\hat{y}_t)},~~~
        \hat{y}_t=\arg\min_k 1-\frac{f_t(x_t)\cdot \mu_k^{(1)}}{||f_t(x_t)||_2||\mu_k^{(1)}||_2},
\end{equation}
where $\xi(*)$ is an indicator that produces 1 when the argument is true. Although the pseudo-labels and the centroids can be updated by Eq.~\eqref{eq:loop} multiple times, we find that one round of updating is sufficient. Given the generated pseudo-labels, the loss function for self-labeling is calculated using the cross-entropy loss, formulated by:
\begin{equation}\label{eq:ce target}
    \mathcal{L}_{sl}=-\mathbb{E}_{x_t\in X_t}\sum_{k=1}^K\xi(\hat{y}_t=k)\log\sigma(\phi_t(x_t)).
\end{equation}

\textbf{Self-Knowledge Distillation}. Recall that we aim to encourage the network to focus on the objects in order to produce more robust feature representations. Although we inject a Transformer module into the model to achieve this goal, we hope to further improve object attention ability by learning with the target samples. The above loss functions ($\mathcal{L}_{im}$ and $\mathcal{L}_{sl}$) are designed to align the feature distribution of the domain, but do not explicitly consider the attention constraint. Therefore, they cannot further improve object attention ability of the Transformer. DINO~\cite{caron2021emerging} showed that learning with a self-knowledge distillation strategy can lead the Transformer to capture more semantic information, \ie, pay more attention to objects. Inspired by this observation, we propose to adopt the self-knowledge distillation strategy to force the model to turn more attention to objects in the target samples. 

Specifically, we employ a teacher model $\phi_t^{Tea}$ and a student model $\phi_t^{Stu}$ to implement self-knowledge distillation. We use the teacher model $\phi_t^{Tea}$ to generate pseudo-labels and optimize the parameters of the student model $\phi_t^{Stu}$ with training losses. Hence, Eq.~\eqref{eq:initial center}, Eq.~\eqref{eq:initial pseudo lable}, and Eq.~\eqref{eq:loop} are re-formulated by replacing $f_t$ and $g_t$ with $f_t^{Tea}$ and $g_t^{Tea}$, respectively. Similarly, $\mathcal{L}_{im}$ and $\mathcal{L}_{sl}$ are re-formulated by replacing $\phi_t$ with $\phi_t^{Stu}$.

For the self-knowledge distillation, we generate the soft pseudo-labels by
\begin{equation}
    \bar{y}_t=\frac{{\rm exp}(\delta(f^{Tea}_t(x_t),\mu_k^{(1)})/\tau)}{\sum^K_{k=1} {\rm exp}(\delta(f^{Tea}_t(x_t),\mu_k^{(1)})/\tau},
\end{equation}
where $\delta(a,b)$ indicates the cosine distance between $a$ and $b$. Then, our knowledge distillation loss is formulated as:
\begin{equation}
    \mathcal{L}_{kd}= -\mathbb{E}_{x_t\in X_t}\sum^K_{k=1}\bar{y}_t \log \phi_t^{Stu}(x_t).
\end{equation}

In summary, the final objective of self-training on the target domain is given by
\begin{equation}\label{eq:all}
    \mathcal{L}_{tgt}=\mathcal{L}_{im}+\alpha \mathcal{L}_{sl} +\beta \mathcal{L}_{kd},
\end{equation}
where $\alpha$ and $\beta$ are the weights of self-labeling loss and self-knowledge distillation loss. We update $\phi_t^{Stu}$ with $\mathcal{L}_{tgt}$. Meanwhile, we update $\phi_t^{Tea}$ with an exponential moving average of the parameters of $\phi_t^{Stu}$. Note that the classifiers of $\phi_t^{Tea}$ and $\phi_t^{Stu}$ are both fixed during training.

%% file: tex/experiments.tex
\section{Experiments}\label{sec:experiments}

\begin{table*}[!t] \small
% \footnotesize
\scriptsize
\caption{Accuracy (\%) on Office-31 for closed-set domain adaptation (ResNet-50).}
\label{tab:office-31}
\centering
\resizebox{0.6\linewidth}{!}{% 
\begin{tabular}{lccccccc}
\toprule
Method & A$\to$D & A$\to$W & D$\to$W & W$\to$D & D$\to$A & W$\to$A & Avg\\
\midrule
ETD~\cite{li2020enhanced}  & 88.0& 92.1 & \textbf{100.0} & \textbf{100.0} & 71.0  & 67.8 & 86.2 \\
BDG~\cite{yang2020bi}  & {93.6} & {93.6} &{99.0} & \textbf{100.0} &{73.2} &{72.0} & {88.5}  \\
CDAN+BSP~\cite{chen2019transferability} & 93.0 & 93.3 & 98.7 & \textbf{100.0} & 73.6 & 72.6 & 88.5\\
CDAN+BNM~\cite{cui2020towards} & 92.9 & 92.8 & 98.8 & \textbf{100.0} & 73.5 & 73.8 & 88.6 \\
CDAN+TransNorm~\cite{Wang19TransNorm} & 94.0 & \textbf{95.7} & 98.7  & \textbf{100.0} & 73.4 & 74.2 & 89.3 \\
GVB-GD~\cite{cui2020gradually} & 96.1 & 93.8 & 98.8 & \textbf{100.0} & 74.9 & 72.8 & 89.4 \\
GSDA~\cite{hu2020unsupervised} & 94.8 & \textbf{95.7} & 99.1 & \textbf{100.0} & 73.5 & 74.9 & 89.7 \\
% \midrule
SHOT~\cite{liang2020we} & 94.0 & 90.1 & 98.4 & 99.9  & 74.7 & 74.3 & 88.6\\
3C-GAN~\cite{li2020model} & 92.7 & 93.7 & 98.5 & 99.8  & 75.3 & 77.8 & 89.6 \\
% Source model only & 83.3 & 83.5 & 97.1 & 99.8 & 59.0 & 62.3 & 80.8 \\
% \ours w/o KD & 96.4 & 94.6 & 99.0 & 99.8  & 73.9 & 76.3 & 90.0 \\
CAN~\cite{kang2019contrastive} & 95.0 & 94.5 & 99.1 & 99.8  & \textbf{78.0} & 77.0 & 90.6 \\
TransDA~(Ours) & \textbf{97.2} & 95.0 & 99.3 & 99.6  & 73.7 & \textbf{79.3} & \textbf{90.7}\\
\bottomrule
\end{tabular}}
\vspace{-0.4cm}
\end{table*}

\begin{table*}[!t] \small
\centering
\caption{Accuracy (\%) on Office-Home for closed-set domain adaptation (ResNet-50).}
\label{tab:office-home}
\resizebox{1\linewidth}{!}{% 
\begin{tabular}{lccccccccccccc}
\toprule
Method & Ar$\to$Cl & Ar$\to$Pr & Ar$\to$Rw & Cl$\to$Ar & Cl$\to$Pr & Cl$\to$Rw & Pr$\to$Ar & Pr$\to$Cl & Pr$\to$Rw & Rw$\to$Ar & Rw$\to$Cl & Rw$\to$Pr & Avg  \\
\midrule
ETD~\cite{li2020enhanced} & 51.3  & 71.9  & 85.7  & 57.6  & 69.2  & 73.7  & 57.8  & 51.2  & 79.3  & 70.2  & 57.5  & 82.1  & 67.3 \\
BDG~\cite{yang2020bi} & 51.5 & {73.4} & {78.7} & {65.3} & {71.5} & 73.7 & {65.1} & 49.7 & {81.1} & {74.6} & 55.1 & {84.8} & {68.7}\\
CDAN+BNM~\cite{cui2020towards} & 56.2  & 73.7  & 79.0  & 63.1  & 73.6  & 74.0  & 62.4  & 54.8  & 80.7  & 72.4  & 58.9  & 83.5  & 69.4 \\
CDAN+TransNorm~\cite{Wang19TransNorm} & 56.3  & 74.2  & 79.0  & 63.9  & 73.5  & 73.1  & 62.3  & 55.2  & 80.3  & 73.5  & 58.4  & 83.3  & 69.4 \\
GVB-GD~\cite{cui2020gradually} & 57.0 & 74.7 & 79.8 & 64.6 & 74.1 & 74.6 & 65.2 & 55.1 & 81.0  & 74.6 & 59.7 & 84.3 & 70.4\\
GSDA~\cite{hu2020unsupervised} & 61.3  & 76.1  & 79.4  & 65.4  & 73.3  & 74.3  & 65.0  & 53.2  & 80.0  & 72.2  & 60.6  & 83.1  & 70.3 \\
RSDA+DANN~\cite{gu2020spherical} & 53.2  & 77.7  & 81.3  & 66.4  & 74.0  & 76.5  & 67.9  & 53.0  & 82.0  & 75.8  & 57.8  & 85.4  & 70.9 \\
% \midrule
SHOT~\cite{liang2020we} & 57.1  & 78.1  & 81.5  & 68.0  & 78.2  & 78.1  & 67.4  & 54.9  & 82.2  & 73.3  & 58.8  & 84.3  & 71.8 \\
% Source model only & 51.7 & 71.6	& 75.2 & 61.9 & 71.5 & 71.2 & 61.9 & 50.3 & 74.7 & 66.9 & 54.4 & 77.2 & 65.7 \\
% \ours w/o KD & 67.2  & \textbf{83.4}  & 85.3  & \textbf{74.4}  & 80.7  & 82.8  & 76.9  & \textbf{68.6}  & 86.9  & 80.0  & \textbf{69.9}  & 89.6  & 78.8 \\
TransDA~(Ours) & \textbf{67.5}  &\textbf{83.3}   & \textbf{85.9}  & \textbf{74.0}  & \textbf{83.8}  & \textbf{84.4}  & \textbf{77.0}  & \textbf{68.0}  & \textbf{87.0}  & \textbf{80.5}  & \textbf{69.9}  & \textbf{90.0}  & \textbf{79.3}\\
\bottomrule
\end{tabular}}
\vspace{-0.4cm}
\end{table*}

\subsection{Experimental Setup}

\textbf{Datasets}. We conduct experiments on three datasets, including Office-31~\cite{saenko2010adapting}, Office-Home~\cite{venkateswara2017deep}, and VisDA~\cite{peng2017visda}. Office-31 includes 4,652 images and 31 categories from three domains, \ie, Amazon (A), Webcam (W), and DSLR (D). Office-Home consists of around 15,500 images from 65 categories. It is composed of four domains: Artistic images (Ar),  Clip Art (Cl), Product images (Pr), and Real-World images (Rw). VisDA contains 152K synthetic images (regarded as the source domain) and 55K real object images (regarded as the target domain), which are divided into 12 shared classes.

\textbf{Evaluation Settings.} We evaluate the proposed method on three DA settings, including closed-set DA, partial-set DA~\cite{cao2019learning}, and open-set DA~\cite{liu2019separate}. Closed-set DA is a standard setting, which assumes that the source and target domains share the same class set. Partial-set DA assumes that the target domain belongs to a sub-class set of the source domain. In contrast, open-set DA assumes that the target domain includes unknown classes that are absent in the source domain. For closed-set DA, we evaluate our method on all three datasets. For partial-set and open-set DA, we evaluate our method on Office-Home. Following~\cite{liang2020we}, for partial-set DA, we choose 25 classes for the target domain, while all 65 classes are use for the source domain. For open-set DA, we select 25 classes as the shared classes while the other classes make up the unknown class in the target domain.

\textbf{Implementation Details}.\label{sec:details} We use a ResNet-50~\cite{he2016deep} pretrained on ImageNet~\cite{he2016deep} as the feature extractor backbone. Moreover, the Transformer~\cite{dosovitskiy2020image} layers are injected after the backbone, followed by a bottleneck layer with batch normalization~\cite{ioffe2015batch} and a task-specific classifier layer. Different from \cite{liang2020we,ahmed2021unsupervised}, we adopt a teacher-student structure for target adaptation. We use stochastic gradient descent(SGD) with momentum 0.9 and weight decay $10^{-3}$ to update the network. The learning rates are set to $10^{-3}$ for the backbone and Transformer layers and set to $10^{-2}$ for the bottleneck and classifier layers.
For source training, we train the model over 100, 50, and 10 epochs for Office-31, Office-Home, and VisDA, respectively. For target adaptation, the number of epochs is set to 15 for all settings. We set $\alpha$ and $\beta$ in Eq.~\eqref{eq:all} to 0.3 and 1, respectively, which yields consistently high performance across all settings. The batch size is set to 64, and the size of the input image is reshaped to $224{\times} 224$.

\subsection{Comparison with State-of-the-Art Methods}
We first compare the proposed \ours with state-of-the-art methods under the closed-set, partial-set, and open-set DA. For closed-set DA, the compared methods include: ETD~\cite{li2020enhanced}, BDG~\cite{yang2020bi}, BSP~\cite{chen2019transferability}, BNM~\cite{cui2020towards}, TransNorm~\cite{Wang19TransNorm},
GVB-GD~\cite{cui2020gradually}, GSDA~\cite{hu2020unsupervised}, CAN~\cite{kang2019contrastive}, SHOT~\cite{liang2020we}, GSDA~\cite{hu2020unsupervised}, RSDA~\cite{gu2020spherical}, DANN~\cite{ganin2016domain}, CDAN~\cite{long2017conditional}, MDD~\cite{zhang2019bridging}, and 3C-GAN~\cite{li2020model}. For partial-set DA and open-set DA, we compare with ETN~\cite{cao2019learning}, SAFN~\cite{xu2019larger}, SHOT~\cite{liang2020we}, TIM~\cite{kundu2020towards}, OSBP~\cite{saito2018open}, STA ~\cite{liu2019separate}, and BA$^3$US~\cite{liang2020balanced}. In these methods, only SHOT~\cite{liang2020we} and 3C-GAN~\cite{li2020model} are designed for source-free domain adaptation.

\textbf{Results on Closed-Set DA}. We report the results on Office-31, Office-Home, and VisDA in Table~\ref{tab:office-31}, Table~\ref{tab:office-home}, and Table~\ref{tab:visda}, respectively. We can make the following three observations. (1) Our \ours outperforms all compared methods on all datasets, yielding state-of-the-art accuracies for closed-set DA. (2) When using the same backbone, our \ours surpasses the source-free method (SHOT~\cite{liang2020we}) by a large margin. Specifically, when using ResNet-50 as the backbone, \ours outperforms SHOT~\cite{liang2020we} by 2.1\%, 7.5\%, and 8.0\% on Office-31, Office-Home, and VisDA, respectively. This verifies the effectiveness of the proposed \ours for source-free DA. (3) On VisDA, \ours with ResNet-50 can produce competitive results compared to the methods that use ResNet-101 as the backbone, further demonstrating the superiority of the proposed \ours. 

\textbf{Results on Partial-Set and Open-Set DA}. To verify the versatility of \ours, we evaluate it on two more challenging tasks, \ie, partial-set DA and open-set DA. Results on Office-Home are reported in Table~\ref{tab:office-home more}. The advantage of \ours is similar to that for closed-set DA. That is, \ours clearly outperforms the compared methods on both settings. Specifically, \ours is 2.0\% and 4.0\% better than SHOT~\cite{liang2020we} on partial-set DA and open-set DA, respectively. This demonstrates that our Transformer-based structure is effective under various domain adaptation settings.

\begin{table*}[t] \small
\caption{Accuracy (\%) on VisDA for closed-set domain adaptation (ResNet-50). $*$ indicates the methods that use ResNet-101 as the backbone. ${\dag}$ indicates reproduction using the official code.}
\centering
\label{tab:visda}
\resizebox{1\linewidth}{!}{% 
\begin{tabular}{lccccccccccccc}
\toprule
% Method & \rotatebox{90}{airplane} & \rotatebox{90}{bicycle} & \rotatebox{90}{bus}  & \rotatebox{90}{car}  & \rotatebox{90}{horse} & \rotatebox{90}{knife} & \rotatebox{90}{motorcycle} & \rotatebox{90}{person} & \rotatebox{90}{plant} & \rotatebox{90}{skateboard} & \rotatebox{90}{train} & \rotatebox{90}{truck} & \rotatebox{90}{Avg} \\
Method & {airplane} & {bicycle} & {bus}  & {car}  & {horse} & {knife} & {motorcycle} & {person} & {plant} & {skateboard} & {train} & {truck} & {Avg} \\
\midrule
DANN~\cite{ganin2016domain} & - & - & - & - & - & - & - & -  & - & - & - & - & 63.7 \\
CDAN~\cite{long2017conditional} & - & - & - & - & - & - & - & - & - & - & - & - & 70.0    \\
MDD~\cite{zhang2019bridging} & - & - & - & - & - & - & - & - & - & - & - & - & 74.6    \\
RSDA~\cite{gu2020spherical} & - & - & - & - & - & - & - & - & - & - & - & - & 75.8    \\
GSDA~\cite{hu2020unsupervised} & 93.1 & 67.8 & {83.1} & \textbf{83.4} & 94.7 & {93.4}  & {93.4} & {79.5} & {93.0} & 88.8 & 83.4  & 36.7  & 81.5 \\
SHOT~\cite{liang2020we}$^{\dag}$ & 94.5 & 85.7 & 77.3 & 52.2 & 91.6 & 15.7 &	82.6 & 80.3 & 87.8 & 88.0 & 85.1 &	58.8 & 75.0\\
TransDA~(Ours) & \textbf{97.2}  & \textbf{91.1} & 81.0 & 57.5 & \textbf{95.3}  & 93.3  & 82.7 & 67.2 & 92.0  & \textbf{91.8} & \textbf{92.5} & \textbf{54.7} & \textbf{83.0}   \\
\hline
SWD~\cite{lee2019sliced}$*$ & 90.8 & 82.5 & 81.7 & 70.5 & 91.7 & 69.5 & 86.3 & 77.5 & 87.4 & 63.6 & 85.6 & 29.2 & 76.4\\
3C-GAN~\cite{li2020model}$*$ & 94.8 & 73.4 & 68.8 & 74.8 & 93.1 & \textbf{95.4} & 88.6 & \textbf{84.7} & 89.1 & {84.7} & 83.5 & 48.1 & 81.6\\
STAR~\cite{lu2020stochastic}$*$ & 95.0 & 84.0 & \textbf{84.6} & 73.0 & 91.6 & 91.8 & 85.9 & 78.4 & \textbf{94.4} & 84.7 & 87.0 & 42.2 & 82.7\\
SHOT~\cite{liang2020we}$*$ & 94.3 & 88.5 & 80.1 & 57.3 & 93.1 & 94.9 & 80.7 & 80.3 & 91.5 & 89.1 & 86.3 & 58.2 & 82.9\\
% \midrule
% Source model only & 54.7 & 51.2 & 45.5 & 32.3 & 53.6 & 52.4 & 46.5 & 37.8 & 51.7 & 51.6 & 52.0 & 30.7 & 46.7 \\
% \ours w/o KD  & 96.5 & \textbf{93.4} & \textbf{86.5} & 61.7 & 95.1  & \textbf{93.5}  & 73.8 & 69.8 & 91.5  & 91.5 & 88.9  & 30.1 & 81.0  \\
%
\bottomrule
\end{tabular}}
\vspace{-0.4cm}
\end{table*}

\begin{table*}[!t] \small
\caption{Accuracy (\%) on Office-Home for partial-set and open-set domain adaptation (ResNet-50).}
\centering
\label{tab:office-home more}
\resizebox{1\linewidth}{!}{% 
\begin{tabular}{lccccccccccccc}
\toprule
Partial-set DA    & Ar$\to$Cl & Ar$\to$Pr & Ar$\to$Rw & Cl$\to$Ar & Cl$\to$Pr & Cl$\to$Rw & P$\to$Ar & Pr$\to$Cl & Pr$\to$Rw & Rw$\to$Ar & Rw$\to$Cl & Rw$\to$Pr & Avg  \\
\midrule
IWAN~\cite{zhang2018importance} & 53.9 & 54.5 & 78.1 & 61.3 & 48.0 & 63.3 & 54.2 & 52.0 & 81.3 & 76.5 & 56.8 & 82.9 & 63.6\\
SAN~\cite{cao2018partial} & 44.4 & 68.7 & 74.6 & 67.5 & 65.0 & 77.8 & 59.8 & 44.7 & 80.1 & 72.2 & 50.2 & 78.7 & 65.3\\
ETN~\cite{cao2019learning} & 59.2 & 77.0  & 79.5  & 62.9  & 65.7  & 75.0  & 68.3  & 55.4  & 84.4  & 75.7  & 57.7  & 84.5  & 70.5 \\
SAFN~\cite{xu2019larger}  & 58.9  & 76.3  & 81.4  & 70.4  & 73.0  & 77.8  & 72.4  & 55.3  & 80.4  & 75.8  & 60.4  & 79.9  & 71.8 \\
BA$^3$US~\cite{liang2020balanced}   & 60.6  & 83.2  & 88.4  & 71.8  & 72.8  & 83.4  & 75.5  & 61.6  & 86.5  & 79.3  & 62.8  & 86.1  & 76.0 \\
SHOT~\cite{liang2020we}   & 64.8  & \textbf{85.2}  & \textbf{92.7}  & \textbf{76.3}  & 77.6  & \textbf{88.8}  & 79.7  & 64.3  & \textbf{89.5}  & 80.6  & 66.4  & 85.8  & 79.3 \\
% \midrule
% Source model only & 57.8 & 63.0 & 72.0 & 57.0 & 66.1 & 68.1 & 64.2 & 56.2 & 68.8 & 69.5 & 59.3 & 70.7 & 64.4 \\
% \ours w/o KD & 71.7  & 81.3  & 89.5  & 74.5  & 85.0  & 83.4  & 82.2  & 68.9  & 87.5  & 87.7  & 70.9  & 88.1  & 80.9 \\
TransDA~(Ours) & \textbf{73.0}  & 79.5  & 90.9  & 72.0  & \textbf{83.4}  & 86.0  & \textbf{81.1}  & \textbf{71.0}  & 86.9  & \textbf{87.8}  & \textbf{74.9}  & \textbf{89.2}  & \textbf{81.3} \\
\midrule
\midrule
Open-set DA & Ar$\to$Cl & Ar$\to$Pr & Ar$\to$ & Cl$\to$Ar & Cl$\to$Pr & Cl$\to$Rw & Pr$\to$Ar & Pr$\to$Cl & Pr$\to$Rw & Rw$\to$Ar & Rw$\to$Cl & Rw$\to$Pr & Avg  \\
\midrule
ATI-$\lambda$~\cite{panareda2017open} & 55.2 & 52.6 & 53.5 & 69.1 & 63.5 & 74.1 & 61.7 & 64.5 & 70.7 & 79.2 & 72.9 & 75.8 & 66.1\\
OSBP~\cite{saito2018open}& 56.7  & 51.5  & 49.2  & 67.5  & 65.5  & 74.0 & 62.5  & 64.8  & 69.3  & 80.6  & 74.7  & 71.5  & 65.7 \\
STA ~\cite{liu2019separate} & 58.1  & 53.1  & 54.4  & \textbf{71.6} & 69.3  & 81.9  & 63.4  & 65.2  & 74.9  & \textbf{85.0} & \textbf{75.8} & 80.8  & 69.5 \\
TIM~\cite{kundu2020towards} &60.1 & 54.2 & 56.2 & 70.9 & 70.0 & 78.6 & 64.0 & 66.1 &  74.9 & 83.2 & 75.7 & 81.3 & 69.6\\
SHOT ~\cite{liang2020we} & 64.5  & \textbf{80.4}  & \textbf{84.7}  & 63.1  & 75.4  & 81.2  & 65.3  & 59.3  & \textbf{83.3}  & 69.6  & 64.6  & 82.3  & 72.8 \\
% \midrule
% Source model only & 46.0 & 50.7 & 54.0 & 45.7 & 49.3 & 52.5 & 43.6 & 43.2 & 53.2 & 48.7 & 46.7 & 56.0 & 49.1 \\
% \ours w/o KD & 71.9  & 79.1  & 84.3  & 71.4  & 77.1  & 82.0  & 68.2  & 67.5  & 83.1  & 76.0  & 73.0  & 87.6  & 76.8 \\
TransDA~(Ours) & \textbf{71.9} & 79.1 & 84.3 & 71.4 & \textbf{77.1} & \textbf{82.0} & \textbf{68.2} & \textbf{67.5} & 83.1 & 76.0 & 73.0 & \textbf{87.6} & \textbf{76.8}\\
\bottomrule
\end{tabular}}
\vspace{-0.4cm}
\end{table*}

\subsection{Ablation Study}

\textbf{Accuracy Comparison}. In Table~\ref{tab:ab-con}, we study the effectiveness of the proposed Transformer structure and self-knowledge distillation. We first explain the components in Table~\ref{tab:ab-con}: \emph{Source Only} indicates the pretrained source model; \emph{Baseline} denotes further training the model on the target data with the information maximization and self-labeling losses; \emph{+Transformer} means injecting the Transformer module into the model; \emph{+EMA} refers to using the teacher-student structure; and \emph{+KD} indicates using the self-knowledge distillation loss. From Table~\ref{tab:ab-con}, we can draw the following conclusions. (1) Injecting the Transformer into the network can consistently improve the results, regardless the model as learnt on the source data or the target data. Specifically, when using the Transformer, the accuracy of \emph{Baseline} improves from 72.1\% to 78.8\% on Office-Home. This demonstrates the effectiveness of the Transformer in domain adaptation. (2) Adding the knowledge distillation loss can further boost the performance, verifying the advantage of the self-knowledge distillation. (3) Applying the teacher-student structure fails to produce clear improvements, indicating that the gains of \emph{+KD} are mainly obtained by the knowledge distillation rather than by generating pseudo-labels with the teacher model.

\begin{figure*}[t] \small
        \centering
    \includegraphics[width=0.95\linewidth]{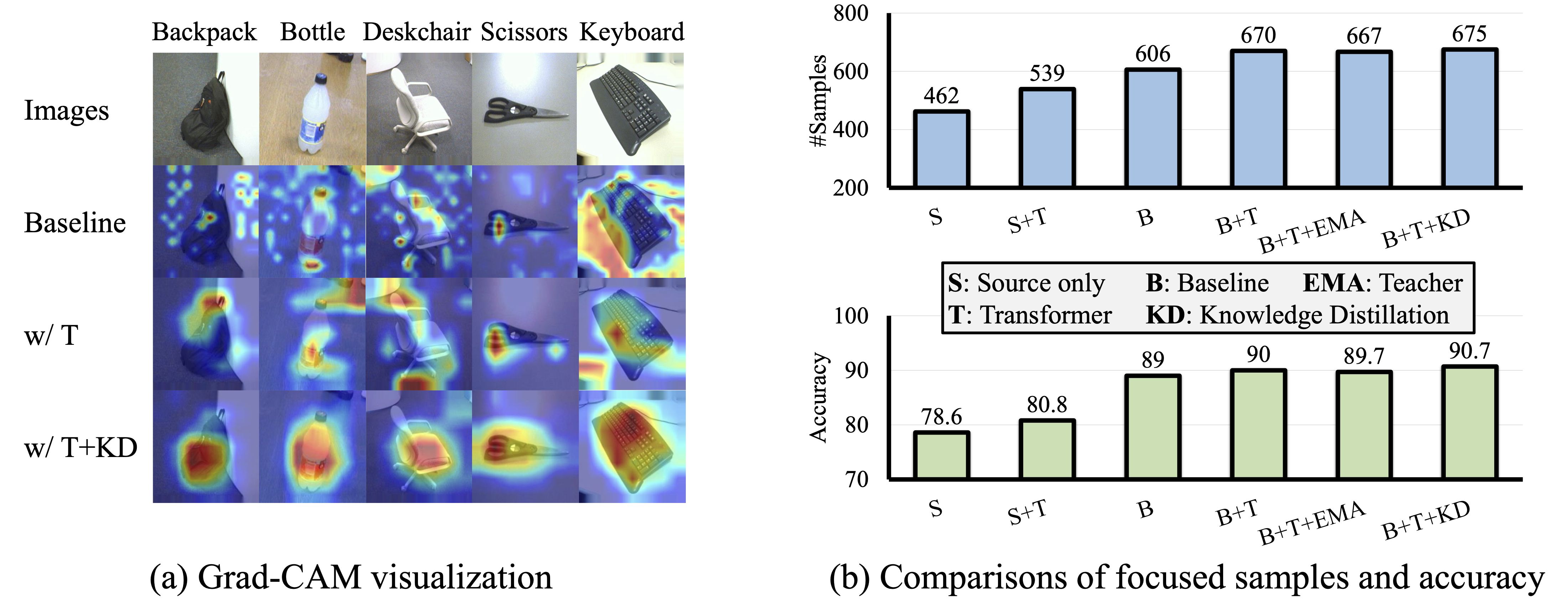}
    \vspace{-.05in}
    \caption{Visualization of (a) Grad-CAM and (b) statistics studies for different methods. Results are evaluated on Office-31 (A$\to$C).}
    \label{fig:vis-ab}
    \vspace{-0.4cm}
\end{figure*}

\textbf{Visualization of Grad-CAM and Statistics Study for Focused and Non-Focused Samples}.
In Figure~\ref{fig:vis-ab}(a), we compare the Grad-CAM visualizations for different variants of our method. We obtain the following findings. (1) When adding the Transformer into the network, the red regions on the objects increase, indicating that the network is encouraged to pay more attentions on the objects. (2) When training the model with the knowledge distillation loss, the attention ability of the network is further improved. In addition, we use Amazon Mechanical Turk to estimate the ``focused / non-focused'' samples for different methods (Figure~\ref{fig:vis-ab}(b)). We can observe that the focused samples and accuracies increase when adding the Transformer and the knowledge distillation loss. The above observations verify that 1) the proposed Transformer structure and knowledge distillation loss can effectively encourage the network to focus on objects, and 2) improving the attention ability of the network can consistently improve the accuracy for domain adaptation.

\begin{figure*}[h] \small
\centering  
\subfigure[Baseline]{
\label{a}
\includegraphics[width=0.3\textwidth,height=1.2in]{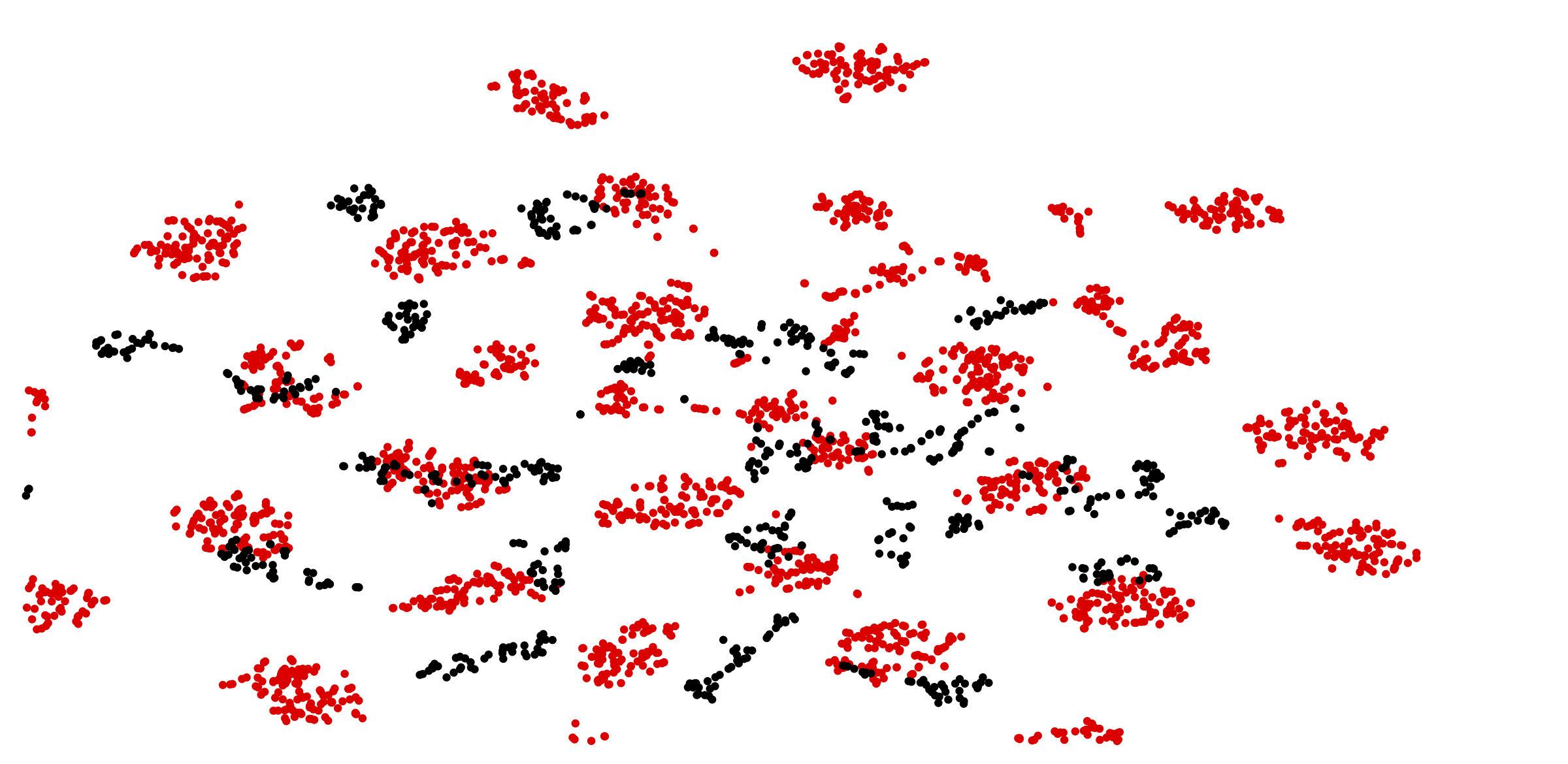}}
\subfigure[Baseline+Transformer]{
\label{b}
\includegraphics[width=0.3\textwidth,height=1.2in]{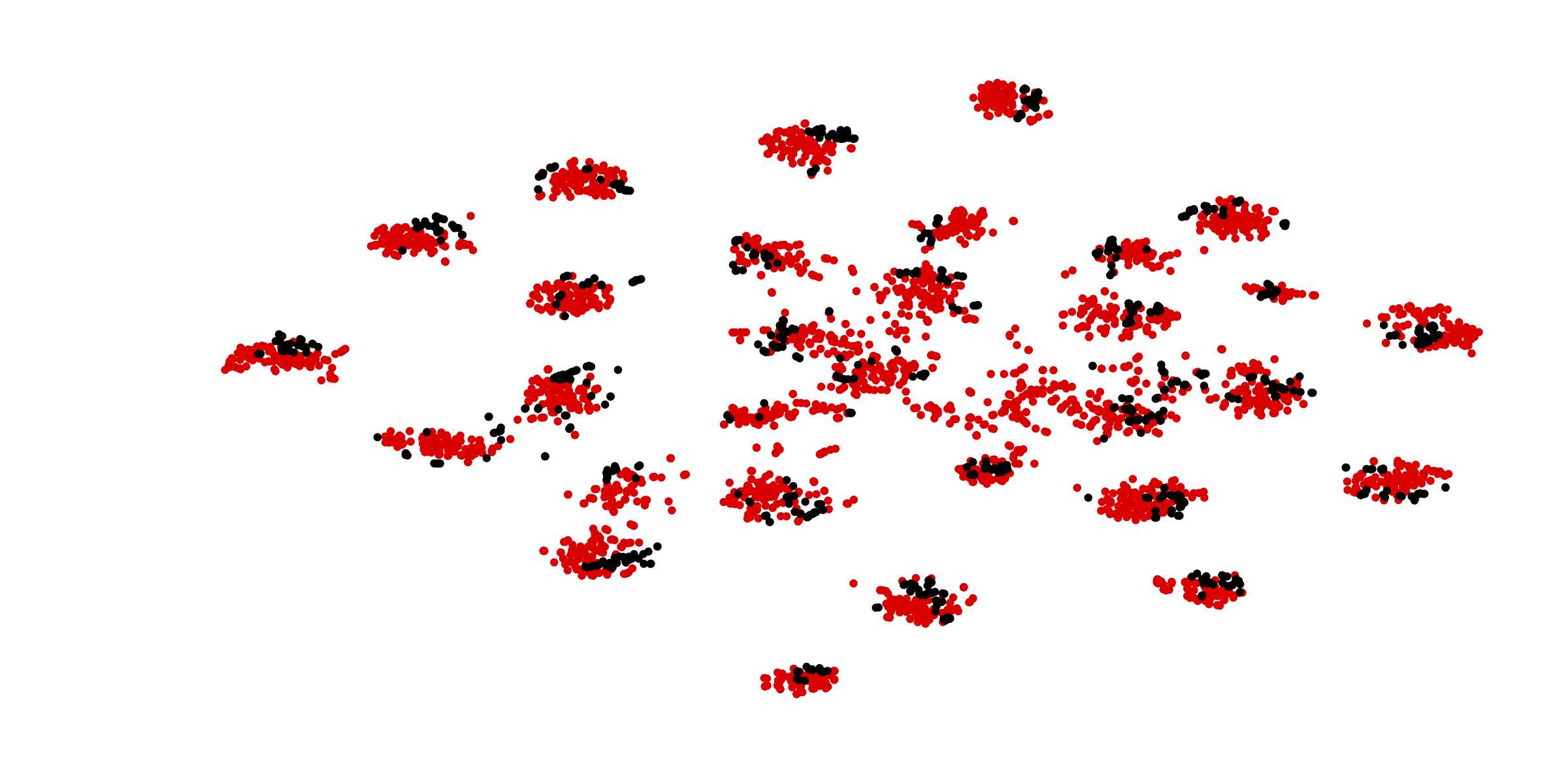}}
\subfigure[Baseline+Transformer+KD]{
\label{v}
\includegraphics[width=0.3\textwidth,height=1.2in]{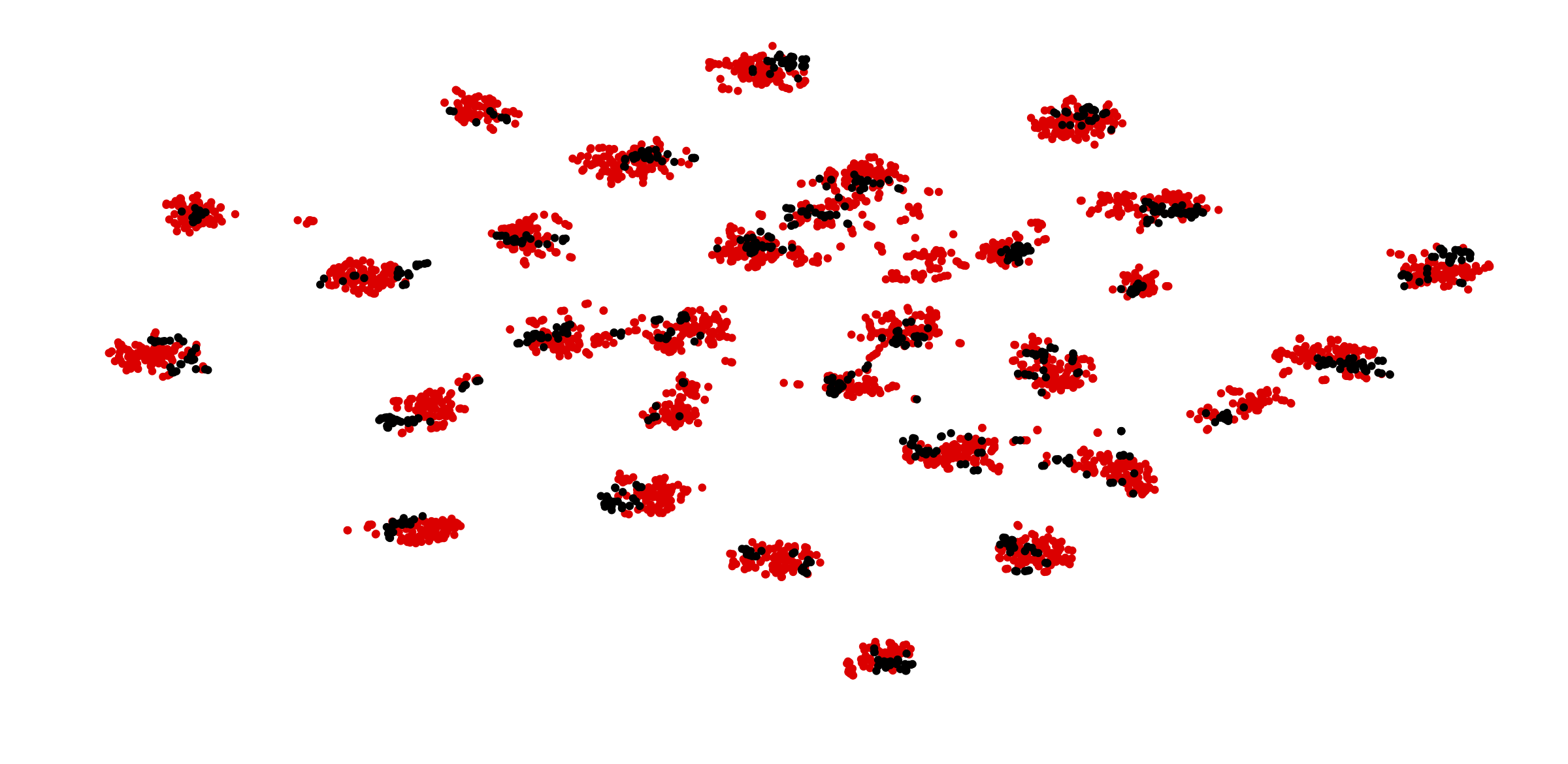}}
\vspace{-.05in}
\caption{t-SNE visualization for different methods on Office-31 (A$\to$W). We use the outputs of the feature extractor as the features. Red/black denote the source/target domains. Best viewed in color.}
\label{fig:t-SNE}
\end{figure*}

\textbf{t-SNE Visualization}. In Figure~\ref{fig:t-SNE}, we show the t-SNE of features for different methods. We find that adding the Transformer and the knowledge distillation can (1) lead the intra-class samples to be more compact and (2) reduce the distances between source and target domains clusters. These findings reveal that \ours can encourage the model to be more robust to intra-class variations and can decrease the cross-domain distribution gap.

\begin{table*}[!t]  \small
\scriptsize
\caption{Ablation study on the Transformer, teacher-student structure (EMA) and self-knowledge distillation (KD). Results are evaluated for the closed-set DA under Office-31, Office-Home, and VisDA.} 
\label{tab:ab-con}
\centering
\begin{tabular}{lccc}
\toprule
Method     & Office-31 & Office-Home & VisDA \\
\midrule
Source Only & 78.6 & 65.7 & 46.7\\
+ Transformer & \bf 80.8 & \bf 67.6 & \bf 48.0\\
\hline
Baseline & 89.0 & 72.1 & 75.0 \\
+ Transformer  & 90.0 & 78.8 & 81.0 \\
+ Transformer + EMA & 90.2 & 78.7 & 81.2 \\
+ Transformer + KD & \bf 90.7 & \bf 79.3 & \bf 83.0 \\
\bottomrule
\end{tabular}
\vspace{-0.4cm}
\end{table*}

%% file: tex/conclusion.tex
\section{Conclusion}\label{sec:conclusion}
\vspace{-.1in}
In this paper, we propose a generic yet straightforward representation learning framework, named TransDA, for source-free domain adaptation (SFDA). Specifically, by employing a Transformer module and learning the model with the self-knowledge distillation loss, the network is encouraged to pay more attention to the objects in an image. Experiments on closed-set, partial-set, and open-set DA confirm the effectiveness of the proposed TransDA. Importantly, this work reveals that the attention ability of a network is highly related to its adaptation accuracy. We hope these findings will provide a new perspective for designing domain adaptation algorithms in the future.